\documentclass[conference]{IEEEtran}
\usepackage[utf8]{inputenc} %Acentos
\usepackage[T1]{fontenc}

\usepackage{hyperref}
\usepackage{amsmath}
\usepackage{enumerate}

\usepackage{algorithmic} %Algoritmos
\usepackage{graphicx} %Gráficas
\usepackage{enumitem}
\usepackage{subcaption}

\usepackage{pdfpages}

\usepackage{listings}

\usepackage{xcolor}

\lstdefinestyle{yaml}{
    basicstyle=\color{blue}\footnotesize,
    rulecolor=\color{black},
    string=[s]{'}{'},
    stringstyle=\color{blue},
    comment=[l]{:},
    commentstyle=\color{black},
    morecomment=[l]{-}
}

\begin{document}
    \title{Multilayer occupancy grid for obstacle avoidance in an autonomous ground vehicle using RGB-D camera}
    
    \label{sec:autores}
%\author{
%    \IEEEauthorblockN{Jhair S. Gallego and Ricardo E. Ramirez} \\
%    \IEEEauthorblockA{
%        \textit{Department of Mechanical and Mechatronics Engineering}\\
%        \textit{Universidad Nacional de Colombia}\\
%         \\
%        \{jhgallego, reramirezh\}@unal.edu.co
%        }
%}

\author{\IEEEauthorblockN{1\textsuperscript{st} Jhair S. Gallego}
\IEEEauthorblockA{\textit{Department of Mechanical and Mechatronics Engineering} \\
\textit{Universidad Nacional de Colombia}\\
Bogotá, Colombia \\
jhgallego@unal.edu.co}
\and
\IEEEauthorblockN{2\textsuperscript{nd} Ricardo E. Ramirez}
\IEEEauthorblockA{\textit{Department of Mechanical and Mechatronics Engineering} \\
\textit{Universidad Nacional de Colombia}\\
Bogotá, Colombia \\
reramirezh@unal.edu.co}}
    \maketitle
    
    \begin{abstract}\label{sec:resumen}    
        This work describes the process of integrating a depth camera into the navigation system of a self-driving ground vehicle (SDV) and the implementation of a multilayer costmap that enhances the vehicle's obstacle identification process by expanding its two-dimensional field of view, based on 2D LIDAR, to a three-dimensional perception system using an RGB-D camera. This approach lays the foundation for a robust vision-based navigation and obstacle detection system. A theoretical review is presented and implementation results are discussed for future work.
    \end{abstract}

    \section{Introduction}

\subsection{Problem definition}

\begin{figure}[htpb]
    
    \centering
    \includegraphics[width=0.8\columnwidth]{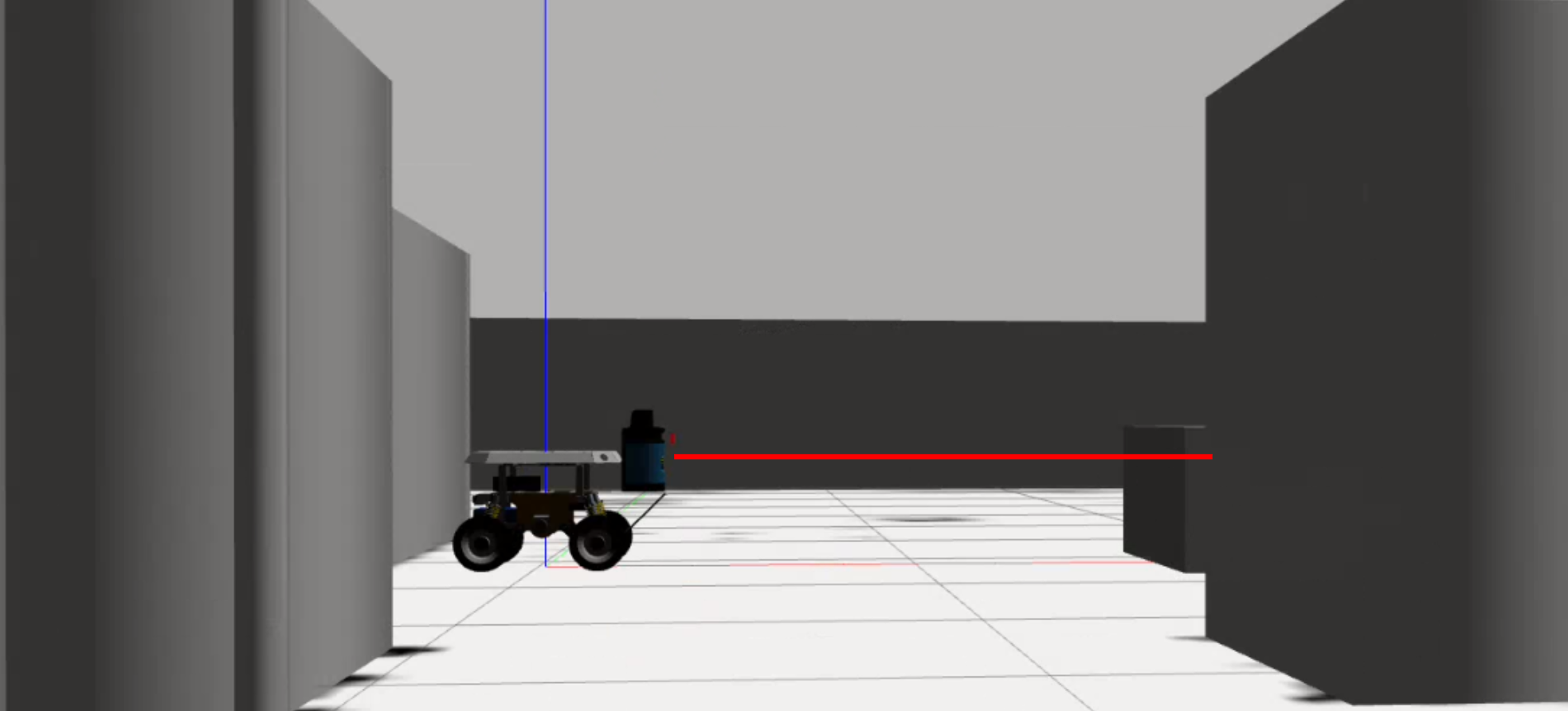}
    \caption{Vertical field of view concept for LiDAR 2D.}
    \label{fig:LiDAR_lateral_vertical_FOVs}
\end{figure}

With the development of new technologies within the context of Industry 4.0, there is a push for intelligent and collaborative communication between robots in a factory. The main objective of this approach is for robots to be flexible in adapting their operations according to the changing demands of the factory. The implementation of collaborative robotics allows these agents to interact with one another, promoting flexibility in their operations. Adaptability in the manufacturing environment is not limited to robot interaction; it also addresses the need to adjust factory operations according to market dynamics. In a context where technologies are emerging rapidly and production needs are evolving, it is crucial for each robot to be flexible within the production chain. This flexibility facilitates an agile response to new demands without the need to incorporate new robots in every case.\\
The logistics of raw materials and manufactured items play a fundamental role in production processes. In the context of an experimental factory, there is an urgent need to transport items between the various machines that make up the facility. Self-Driving ground Vehicles (SDVs) enable the transportation of these items within production plants. These vehicles, upon receiving instructions from an external agent, plan trajectories and navigate autonomously, overcoming dynamic obstacles such as the presence of people or other robots.\\
The use of a 2D LiDAR sensor is common in the perception system of these vehicles. However, in certain scenarios, the effectiveness of trajectory execution is impacted due to the limited field of view provided by the sensor. The 2D LiDAR-based perception system cannot detect obstacles whose position or geometry lies outside the sensor's field of view. As shown in Figure \ref{fig:LiDAR_lateral_vertical_FOVs}, any obstacle whose geometry does not intersect with the red plane will affect the effective generation of trajectories that allow for evasion, as the perception system will not account for it, resulting in a collision. This generates operational failures in the production processes of the factories where the vehicles operate. To address this limitation, we propose the integration of a depth camera to the perception system of these vehicles, extending the field of view to three dimensions and accounting for new obstacle detection improvements. In this work, we describe the process of integration of a depth camera (RGB-D) into the navigation system of an SDV, we present a theoretical review of the concepts around this integration, describe the implementation of a multilayer occupancy grid, and evaluate its effectiveness in strengthening the obstacle detection process.

\subsection{Working vehicle}\label{sec:vehiculo_de_trabajo}
\begin{figure}[htbp]
            \centering 
            \includegraphics[width=0.7\columnwidth]{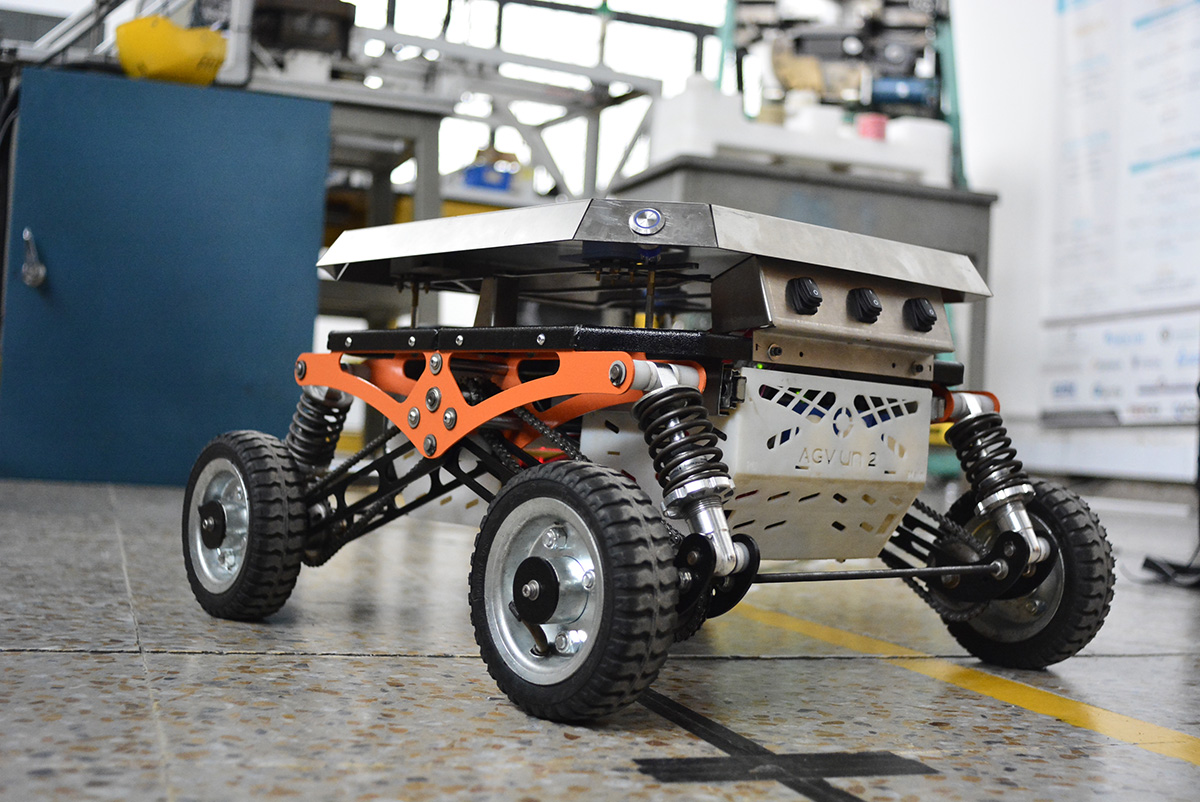}
            \caption{SDV II vehicle from the LabFabEx~\cite{sdvII_unimedios_report}.}\label{fig:sdvII_unimedios}
        \end{figure}

For the development of this work, the SDV II~\cite{agv_2016_UAO} from the Experimental Factory Laboratory (LabFabEx) at the Universidad Nacional de Colombia was used. The vehicle is shown in Figure~\ref{fig:sdvII_unimedios}.

\begin{figure}[htbp]
        \centering
        \includegraphics[width=0.7\columnwidth]{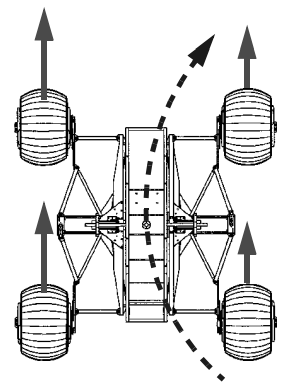}
        \caption{Skid-steering architecture~\cite{skid_steering_diagram}.}\label{fig:skid-steering}
    \end{figure}

        This vehicle features a skid-steering kinematic architecture (see fig.~\ref{fig:skid-steering}), that allows it to move with two degrees of freedom. i.e., in the vehicle's local coordinate frame with origin in the geometric center of the vehicle, movement forward and backwards ($x$ direction) and rotations around the ground plane ($yaw$)~\cite{skid_steering_kinematics}. The vehicle's electronic components are classified as follows:

        \begin{enumerate}
            \item \textit{Computing and Navigation System:} Composed of an Intel NUC mini-computer with 8 GB of RAM, Two USB 2.0 ports, and one USB 3.0 port~\cite{NUC_intel_webpage}. The operating system used is Linux 18.04 (kernel v5.5), and navigation is implemented under the ROS Noetic framework~\cite{ros_noetic_docs}.
        
            \item \textit{Perception System:} Uses the 2D LiDAR sensor Sick-LMS102~\cite{sick_website}.
            
            \item \textit{Low-Level Control:} Uses a Tiva C Series board~\cite{tiva_datasheet}. It facilitates serial communication with the motor drivers responsible for the vehicle's propulsion.
    
    \end{enumerate}

    \section{Background}\label{sec:system_architecture}

This section presents the structure and functioning of specific components of the navigation system (\textit{nav\_stack}) that enable trajectory planning and execution based on data provided by the perception system. The study of the multilayer costmap, used in trajectory generation, is introduced. Finally, the concepts of the operating theory and optical properties of the RGB-D camera are also presented.

\subsection{Navigation System}\label{sec:nodos_nav_stack}

\begin{figure}[htbp]
    \centering
    \includegraphics[width=\columnwidth]{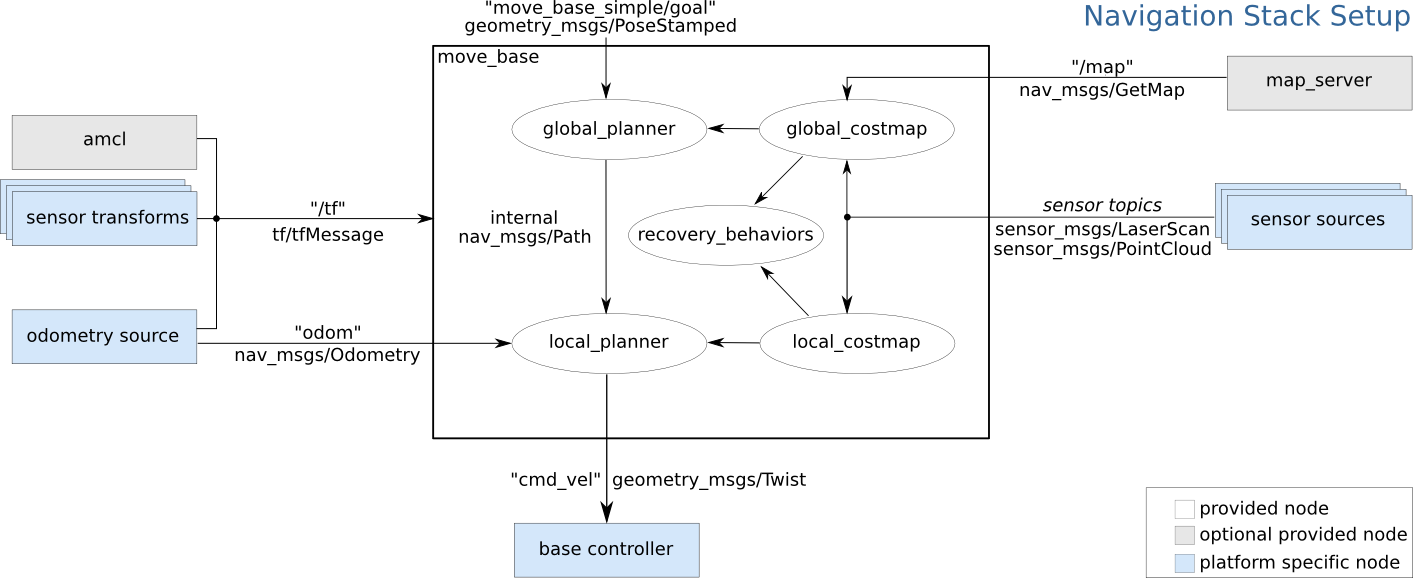}
    \caption{Costmap under the navigation system architecture of the SDV~\cite{move_base_ros_node}.}\label{fig:move_base_arq}
\end{figure}

The autonomous navigation system is implemented using the \textit{move\_base} navigation package from ROS~\cite{move_base_ros_node}. It uses costmaps to represent information about the environment's occupancy by objects. This information is used by local and global planners (\textit{global\_planner} and \textit{local\_planner}) for safe trajectory planning and generation. In the context of ROS's autonomous navigation stack, there are two main levels of costmaps: the \textit{global\_costmap} and the \textit{local\_costmap} (see fig. \ref{fig:move_base_arq}).

\textit{Global Costmap:}  
The global\_costmap aims to provide a high-level occupancy representation of the environment and is used for global route planning. This costmap is generated using both static and dynamic information. The former generally comes from previously created occupancy maps and represents the unchanging environment, such as stationary robots or walls. On the other hand, the dynamic information provides a real-time representation of the robot's environment, such as a moving person or another vehicle, and comes directly from the perception system.

\textit{Local Costmap:}  
The local\_costmap, in contrast, describes the immediate environment of the robot and is used for local navigation. For example, it identifies nearby and moving obstacles. This costmap is updated more frequently and also utilizes information provided by the perception system.

\textit{Occupancy Grid:}  
The main representation of the environment in the global and local costmaps is an occupancy grid. In this grid, each cell represents a region of the environment and has a value indicating the level of occupancy. Typically, a scale of discrete values is used, such as: 0 (free cell), indicating that the cell is unoccupied, or 100 (occupied cell), indicating the presence of an obstacle. This grid is what the trajectory planning system ultimately uses to generate safe routes for navigating the environment.
\subsection{Multilayer Costmap}\label{sec:costmap}

\begin{figure}[htbp]
    \centering
    \includegraphics[width=0.95\columnwidth]{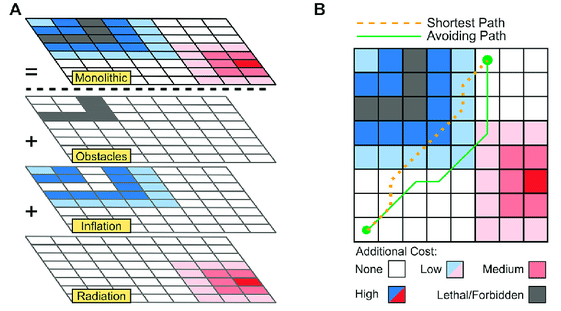}
    \caption{The Multilayer Costmap concept~\cite{costmap_multilayer}.}\label{fig:costmap_multilayer}
\end{figure}

The overlay of the global and local costmaps is referred to as the Multilayer Costmap. The occupancy grid generated from this map integrates both global and local information. Figure \ref{fig:costmap_multilayer} shows a conceptual representation of this map, where each costmap, whether global or local, can originate from specific sensors, such as LiDAR or cameras. Additionally, it is possible to use multiple costmaps from various sensors of the same type, such as several cameras in different positions, thereby strengthening the perception system.

\subsection{Optical Properties of the RGB-D Camera}\label{sec:rgb_d_properties}

    \subsubsection*{Field of View}
\begin{figure}[htbp]
        \centering
        \includegraphics[width=6cm]{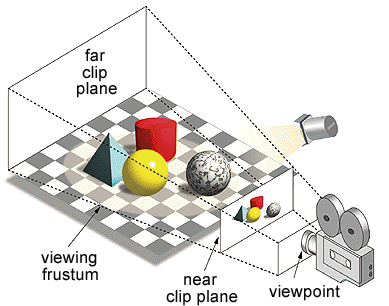}
        \caption{Three-dimensional representation of a camera's field of view~\cite{frustum}.}\label{fig:frustum}
    \end{figure}

    The field of view (FOV) of a camera allows us to understand the extent of the observable scene in an image or video. It represents the angular range that the camera lens can capture. It is generally defined in both horizontal and vertical dimensions. The horizontal field of view determines the left-to-right range, while the vertical field of view covers the up-and-down range. Spatially, the FOV is described by a frustum, which is a geometric shape that plays a crucial role in understanding and visualizing the camera's field of view (see fig.~\ref{fig:frustum}). The frustum defines the volume of space that the camera can capture in a scene and is essentially a truncated pyramid, where the apex of the pyramid is at the camera lens, and the base of the pyramid represents the image plane or sensor.

\begin{figure}[htbp]
        \centering
        \includegraphics[width=\columnwidth]{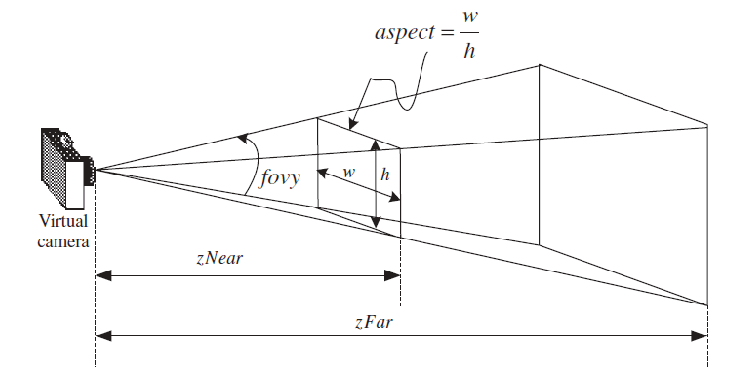}
        \caption{Characteristic parameters of a camera's field of view\cite{frustumfov}.}\label{fig:frustum_fov}
    \end{figure}

    The camera's frustum is defined by three parameters: the near and far clipping planes, the horizontal and vertical fields of view, and the aspect ratio of the image. The near and far clipping planes represent the minimum and maximum distances from the camera at which objects will be in focus and included in the captured image. The shape of the frustum is determined by extending lines from the camera lens through the corners of the near and far clipping planes (see fig. \ref{fig:frustum}). Similarly, the values of the horizontal and vertical FOV (\textit{fovy} in fig. \ref{fig:frustum_fov}), often specified in degrees or radians, indicate the angular extent covered by the lens. Furthermore, the aspect ratio of the image, which represents the proportional relationship between width and height, influences how the FOV is distributed (see fig. \ref{fig:frustum_fov}).

    \subsubsection*{Depth Map}
    The depth map or point cloud is a three-dimensional representation of the environment captured by a depth camera. In this context, depth is measured as the distance from the camera to objects in the scene. The point cloud consists of a set of coordinates ($x, y, z$), where each point represents the spatial position of a detected object in the image captured by the camera.\\
    This type of three-dimensional information is fundamental for perception and autonomous navigation applications in robotics. In the context of ROS, the autonomous navigation stack uses the point cloud to build and update occupancy maps of the robot's environment. These maps are used to plan safe routes and avoid obstacles during autonomous navigation. This type of data is represented using the PointCloud2 message from the ROS framework~\cite{pointCloud2_ros_msg}. The message' structure contains detailed information about the point cloud, including the three-dimensional coordinates of each point, as well as additional information such as data on color and signal intensity.

    \section{Implementation}

\subsection{\textbf{Development Environment}}
\begin{figure}[htbp]
    \centering
    \includegraphics[width=0.95\columnwidth]{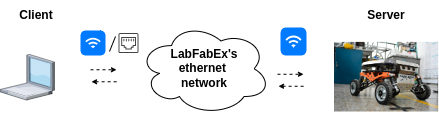}
    \caption{Client-server architecture.}\label{fig:cliente_servidor}
\end{figure}

\begin{figure}[htbp]
    \centering
    \includegraphics[width=0.7\columnwidth]{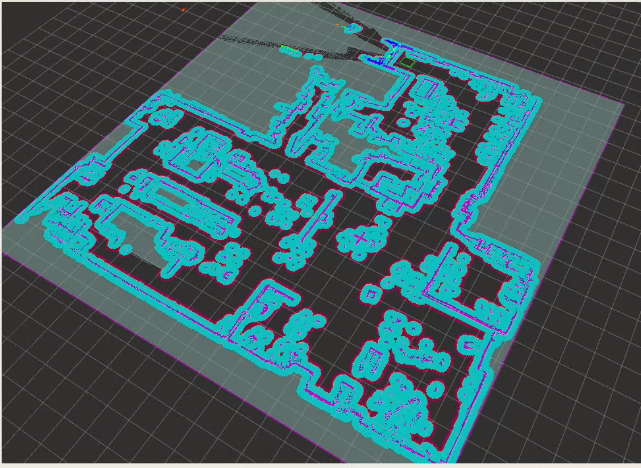}
    \caption{Perception of the surrounding world by the SDV.}\label{fig:rviz_costmap_view}
\end{figure}

%\subsection{Development Environment}
To avoid overloading the operating system on the NUC when running graphical applications on Linux, the distributed architecture of ROS is used under a client-server model (see fig.~\ref{fig:cliente_servidor}). In this way, the NUC can operate without a graphical interface (\textit{headless} mode), and the data published by the topics are consulted over the Wi-Fi network. Visualization and data analysis applications run on a client, a computer connected to the same network, leveraging the resources of both devices. As shown in fig. \ref{fig:rviz_costmap_view}, using the Rviz visualizer from an external computer, we can observe sensors' readings and how it is interpreted by the vehicle. This method facilitates the troubleshooting process and effectively monitors the vehicle's status in real-time. Additionally, this architecture allows multiple clients to query the vehicle's data, improving the scalability of the development environment.

\subsection*{Docker Technology}

To ensure consistency in software versions across different clients and to guarantee the replicability of the development environment, Docker is used. This technology allows the creation of a container defined in a plain text file, commonly called a Dockerfile, which (for the purposes of this work) unifies the use of Rviz, Gazebo, and ROS (Noetic) in a standardized environment, facilitating the replication of the development environment on any client.

\subsection{\textbf{Intel RealSense D435i Depth Camera}}
\begin{figure}[htbp]
    \centering
    \includegraphics[width=\columnwidth]{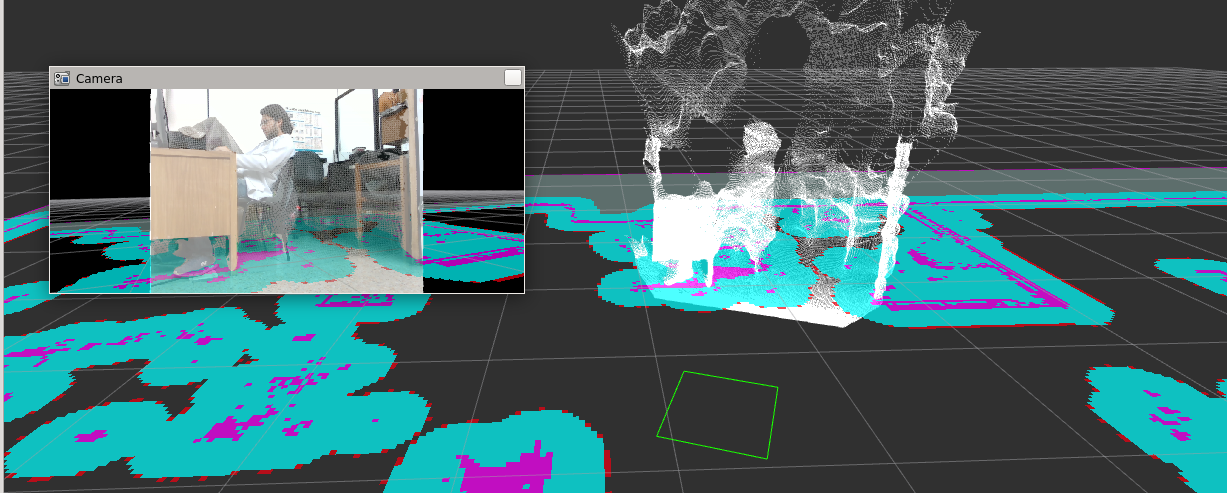}
    \caption{Point cloud generated by the D435i camera at LabFabEx.}\label{fig:d435i_point_cloud}
\end{figure}

%Justification for the change
As a RGB-D camera, the Intel D435i reference is used. This camera offers two main advantages: it can be connected via a USB 2.0 port and has long-term technological support from the manufacturer. For integration with ROS, the installation process recommended by the manufacturer was followed as in \cite{D435i_install_ros}. Figure~\ref{fig:d435i_point_cloud} illustrates the point cloud (in white) and the \textit{stereo} image generated by the camera.

\subsection*{Optical Properties of D435i}\label{sec:D435i_optical_caracteristics}

The technical specifications of the camera relevant to this work are summarized in Table~\ref{table:d435i_fov_specs}.

\begin{table}[htpb]
    \centering
    \caption{D435i FOV Specifications (adapted from \cite{D435i_specs}).}
    \begin{tabular}{|c|c|}
        \hline
        \textbf{Operating Range}        & 0.3m to 3m                       \\
        \hline
        \textbf{Field of View (FOV)}    & 87° (vertical) 58° (horizontal) \\
        \hline
        \textbf{Aspect Ratio}           & 16:9                            \\
        \hline
        \textbf{Depth Resolution}       & 1280 x 720                      \\
        \hline
    \end{tabular}
    \label{table:d435i_fov_specs}
\end{table}

% Section "RGB-D Camera"
\subsection*{Support Design for Mounting the Camera on the Vehicle}
\begin{figure}[htpb]
    \centering
    
    \begin{subfigure}{0.5\columnwidth}
        \centering
        \includegraphics[width=0.9\linewidth]{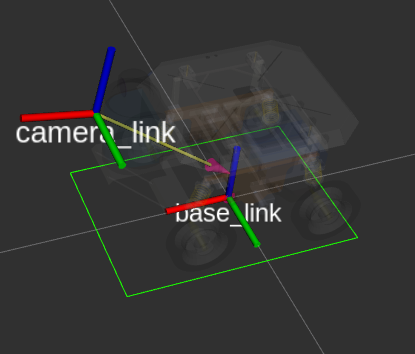}
        \caption{}
        \label{fig:base_to_camera_tf_isometric}
    \end{subfigure}%
    \begin{subfigure}{0.5\columnwidth}
        \centering
        \includegraphics[width=0.9\linewidth]{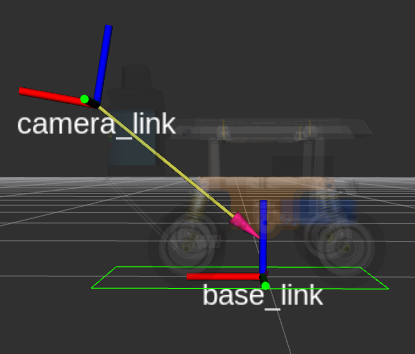}
        \caption{}
        \label{fig:base_to_camera_tf_lateral}
    \end{subfigure}

    \caption{Relationship between the \textit{base\_link} and \textit{camera\_link} coordinate systems.}\label{fig:base_link_to_camera_link}
\end{figure}

The navigation system of the SDV requires knowing the \textit{pose} of the camera in order to perform the coordinate transformation to the central coordinate system of the SDV. This is achieved by providing the location of the camera's geometric center with respect to the central coordinate system (see fig.~\ref{fig:base_link_to_camera_link}). With this, it is possible to properly interpret and process the generated point cloud. Therefore, it is necessary to know the \textit{pose} of the camera to perform the corresponding transformation (see fig.~\ref{fig:base_link_to_camera_link}).

Since it is not necessary for the camera to perform translations or rotations about its own axis, but rather to remain fixed, it is important that the position of the camera relative to the SDV base is always the same. To this end, a support was designed and manufactured to allow the camera to be directly coupled to a pre-existing mechanism on the SDV. This way, the position vector (and hence the coordinate system transformation) of the camera with respect to the base ($TF_{cb}$) can be obtained (see fig.~\ref{fig:base_link_to_camera_link}). The position of the \textit{camera\_link} coordinate system relative to the \textit{base\_link} coordinate system is described by the vector~\ref{eq:tf_cb_vector}:

\begin{equation}
    \mathbf{TF}_{cb} = \langle 0.345, 0, 0.28 \rangle m \label{eq:tf_cb_vector}
\end{equation}

\begin{figure}[htpb]
    \centering
    
    \begin{subfigure}{0.5\columnwidth}
        \centering
        \includegraphics[width=0.8\linewidth]{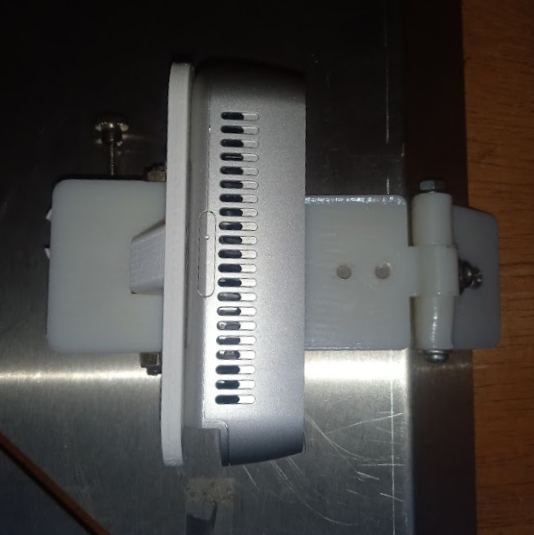}
        \caption{}
        \label{fig:}
    \end{subfigure}%
    \begin{subfigure}{0.5\columnwidth}
        \centering
        \includegraphics[width=\linewidth]{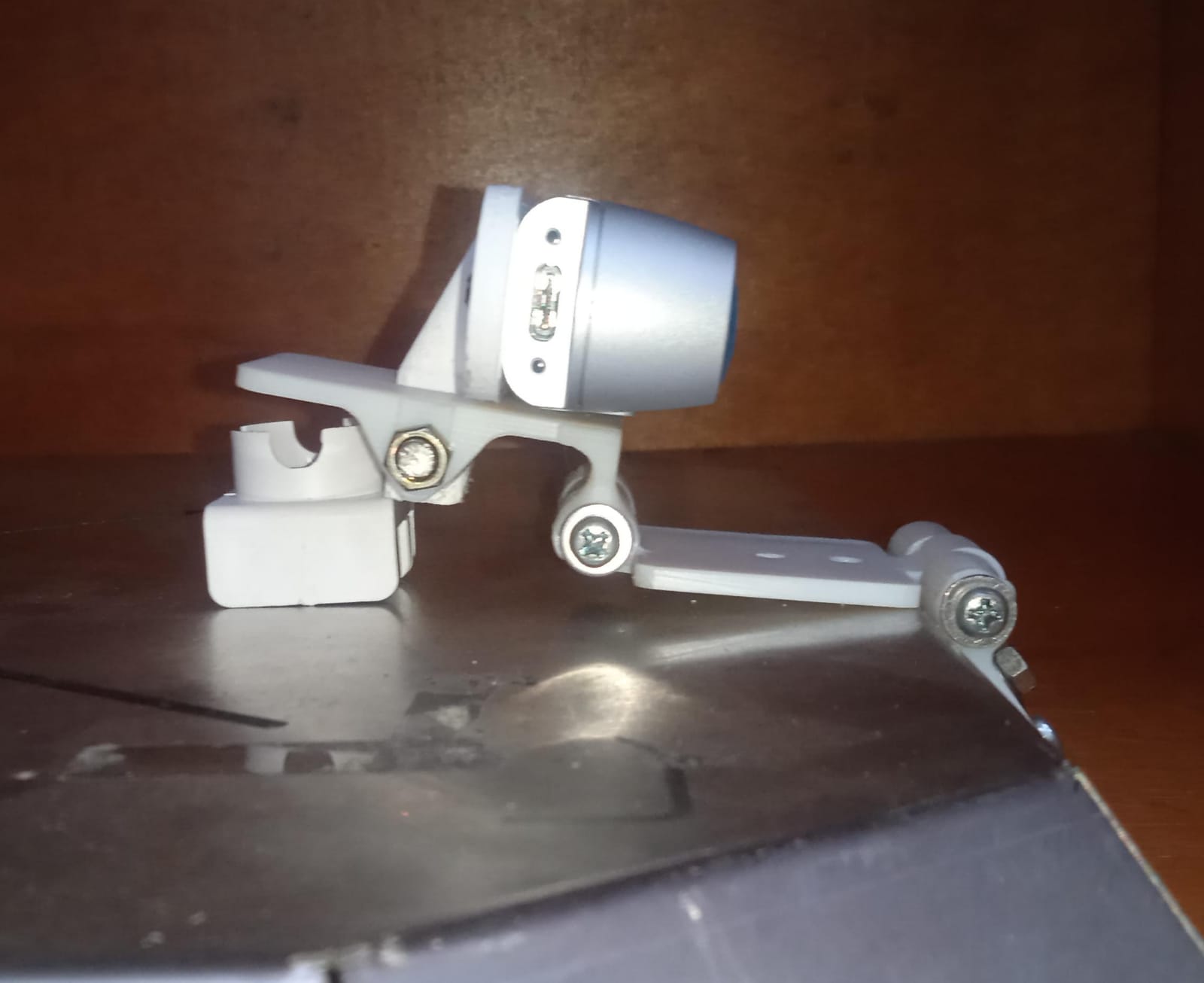}
        \caption{}
        \label{fig:}
    \end{subfigure}
    \hfill
    \begin{subfigure}{0.5\columnwidth}
        \centering
        \includegraphics[width=0.8\linewidth]{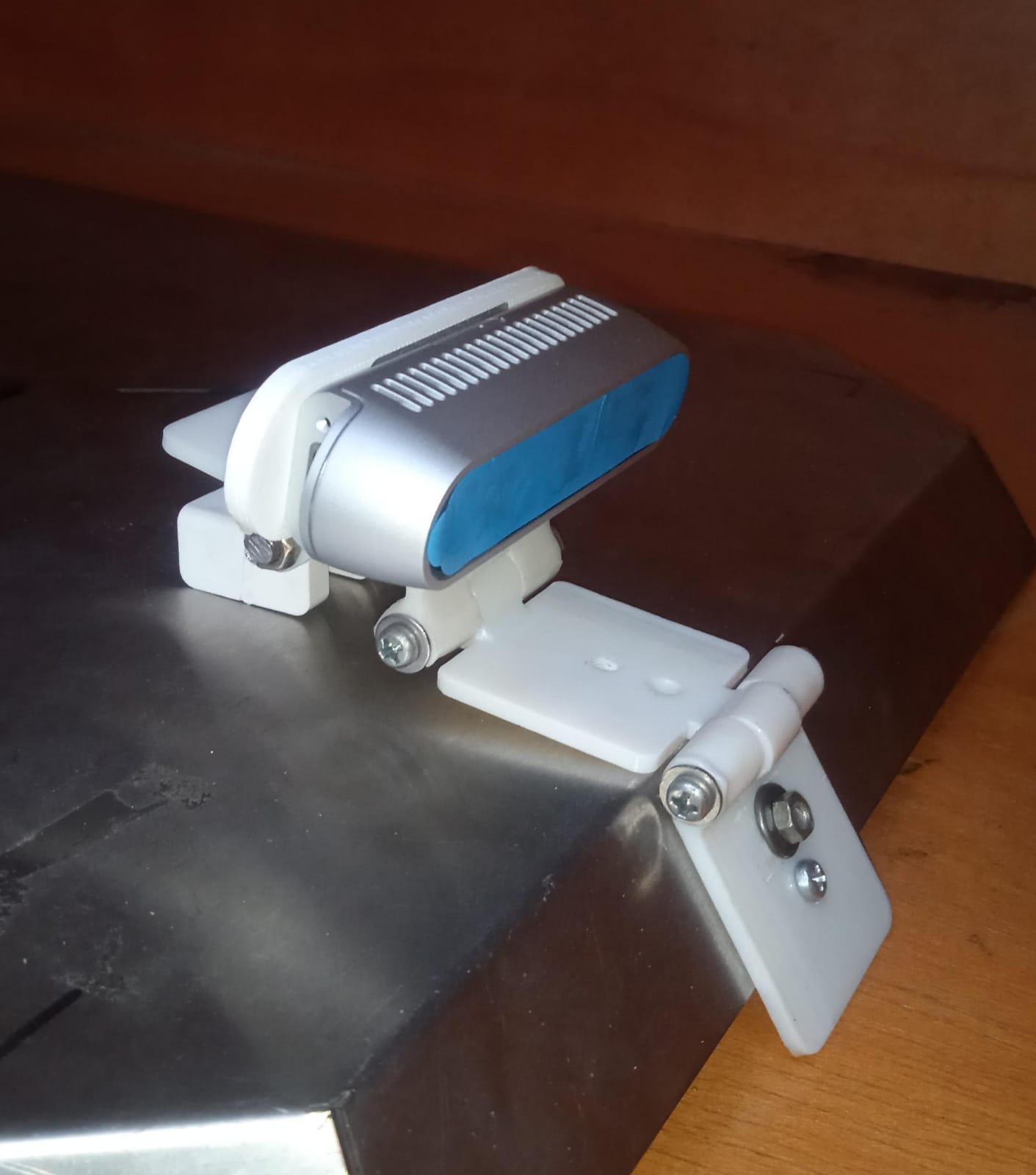}
        \caption{}
        \label{fig:}
    \end{subfigure}%
    \begin{subfigure}{0.5\columnwidth}
        \centering
        \includegraphics[width=\linewidth]{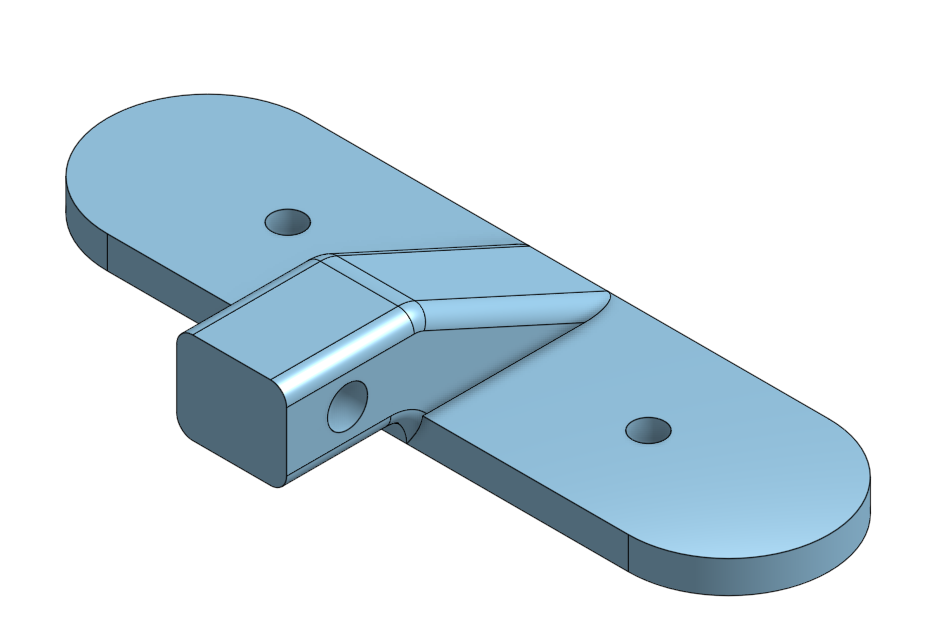}
        \caption{}
        \label{fig:camera_support}
    \end{subfigure}
    
    \caption{D435i camera support. \textit{(a), (b), (c)}: Camera support mounted on mechanism. \textit{(d)}: Isometric view of support's CAD model.}
    \label{fig:mechanism_and_camera_support}
\end{figure}

The final design is shown in fig.~\ref{fig:camera_support}, and the integration with the pre-existing mechanism along with the mounting of the camera is shown in fig.~\ref{fig:mechanism_and_camera_support}. 

\subsection{\textbf{Integration with the Navigation System}}
\begin{figure}[htbp]
    \centering
    \includegraphics[width=\columnwidth]{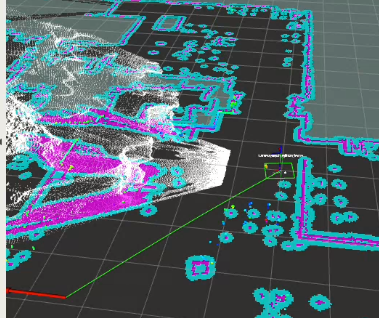}
    \caption{False positive clasification of obstacles when using default detection range limits in \textit{costmap\_2d} parameters.}\label{fig:costmap_pointcloud_ranges}
\end{figure}

%Parameters modified in .yaml files
As mentioned in sections~\ref{sec:nodos_nav_stack} and~\ref{sec:costmap}, integrating a new sensor requires adjusting certain configuration parameters in the navigation system. In Figure~\ref{fig:costmap_pointcloud_ranges}, it can be observed that as the viewing plane of the camera's frustum extends further, the points exhibit greater dispersion and fail to accurately represent objects at those distances, such as walls in this case. Consequently, the costmap generates a false positive for an obstacle (represented in purple), which impacts trajectory planning for target points intersecting the affected areas. Therefore, the detection range of points as obstacles is limited to a frontal distance of less than $2m$ and a height range between $0.35m$ and $1.0m$. Table~\ref{table:costmap_params} provides a summary of these and other relevant configuration parameters. These define a frustum as described in section~\ref{sec:rgb_d_properties} and the ROS topic's name and data type to which the point cloud information comes from. The functionality of each parameter can be referenced from the official documentation of the /move\_base package~\cite{costmap_2d_parameters}.

\begin{table}[htbp]
    \centering
    \caption{Camera integration parameters for costmap}
    \begin{tabular}{|c|c|}
        \hline
        \textbf{data\_type} & PointCloud2  \\
        \hline
        \textbf{topic} & camera/depth/color/points \\
        \hline
        \textbf{marking} & true \\
        \hline
        \textbf{clearing} & true \\
        \hline
        \textbf{max\_obstacle\_height} & 1.0 \\
        \hline
        \textbf{min\_obstacle\_height} & 0.35 \\
        \hline
        \textbf{obstacle\_range} & 2.0 \\
        \hline
        \textbf{raytrace\_range} & 2.0  \\
        \hline
    \end{tabular}
    \label{table:costmap_params}
\end{table}

\section{Results}

\begin{figure}[htbp]
    \centering
    \includegraphics[width=\columnwidth]{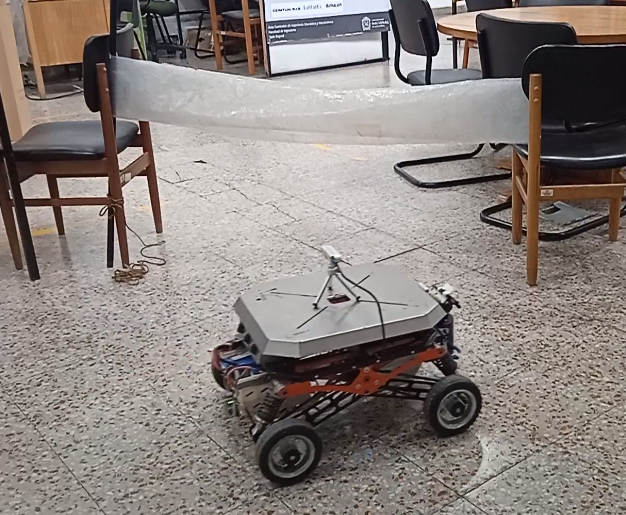}
    \caption{Testing scenario for obstacle avoidance with the RGB-D camera.}\label{fig:plastic_bridge_scenario}
\end{figure}

%\subsection{Verification of obstacle avoidance with the camera}
To verify that the camera is indeed assisting in trajectory planning by identifying objects, the following scenario was implemented. An obstacle in the form of a bridge, represented by plastic (hereafter referred to as \textit{the bridge}), was placed at a height of $60cm$, in such a way that it could not be detected by the LiDAR but could be identified by the camera, due to its vertical field of view (see fig.~\ref{fig:plastic_bridge_scenario}). The test consists of providing a target position to the vehicle, such that it initially generates a trajectory that should pass under the bridge.

\begin{figure}[htpb]
    \centering

    \begin{subfigure}{\columnwidth}
        \centering
        \includegraphics[width=0.9\linewidth]{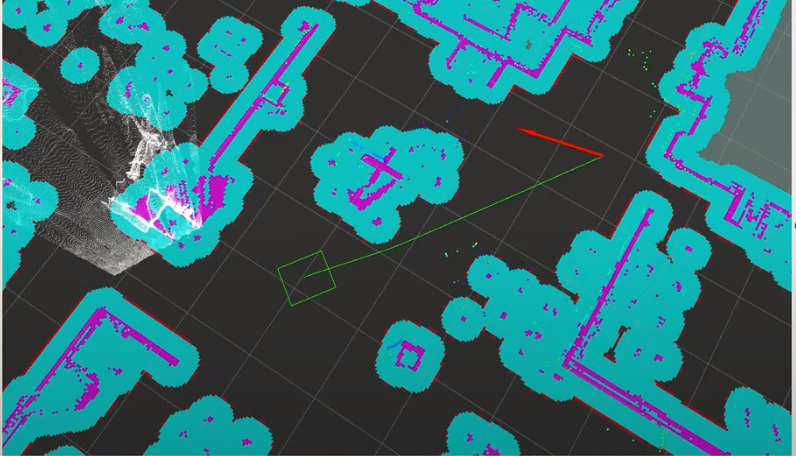}
        \caption{}
        \label{fig:before_plastic_bridge_rviz_view}
    \end{subfigure}
    \hfill
    \begin{subfigure}{\columnwidth}
        \centering
        \includegraphics[width=0.9\linewidth]{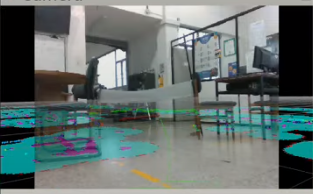}
        \caption{}
        \label{fig:before_plastic_bridge_test_camera_view}
    \end{subfigure}

    \caption{Testing scenario before dynamic obstacle avoidance.}\label{fig:before_plastic_bridge_test}
\end{figure}

\begin{figure}[htpb]
    \centering
    \begin{subfigure}{\columnwidth}
        \centering
        \includegraphics[width=0.9\linewidth]{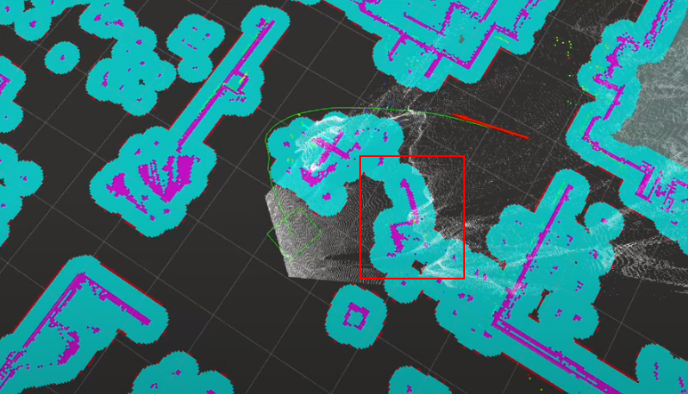}
        \caption{}
        \label{fig:after_plastic_bridge_test_rviz_view_highlighted}
    \end{subfigure}
    \hfill
    \begin{subfigure}{\columnwidth}
        \centering
        \includegraphics[width=0.9\linewidth]{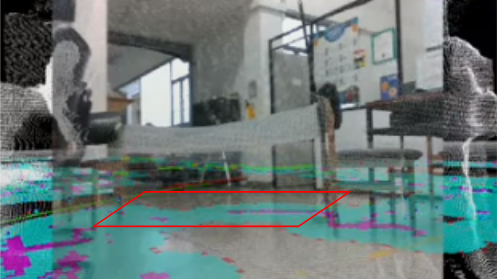}
        \caption{}
        \label{fig:after_plastic_bridge_test_camera_view}
    \end{subfigure}

    \caption{Testing scenario after dynamic obstacle avoidance.}\label{fig:after_plastic_bridge_test}
\end{figure}

In figure~\ref{fig:before_plastic_bridge_test}, it can be seen that initially, a trajectory is generated so that the vehicle should go under the bridge. Once the camera identifies the obstacle, it is effectively projected onto the costmap, highlighted by a red box in figure~\ref{fig:after_plastic_bridge_test_rviz_view_highlighted}. Similarly, an alternative trajectory (green line) is recalculated to reach the proposed target location. In figure~\ref{fig:after_plastic_bridge_test_camera_view}, from the camera's perspective, both the bridge (obstacle) and its projection generated by the navigation system can be observed.

\section{Conclusions}\label{sec:conclusiones}
\begin{itemize}

    \item The integration of the RGB-D camera significantly improves the autonomous vehicle's ability to identify and avoid obstacles, especially those not visible by the LiDAR. This provides robustness in trajectory planning in dynamic environments.
    
    %\item The work addresses previous issues of reliability in motor communication and the limited detection of lateral obstacles. The implementation of the RGB-D camera and the adjustment of costmap parameters, such as the minimum and maximum height of obstacles, reduces detection errors and improves the operational autonomy of the SDV.
    
    \item The use of technologies like Docker to standardize the development environment and ensure experiment replicability is a key point of this work. This allows for efficient integration and testing of new cameras and algorithms, facilitating the maintenance and expansion of the autonomous navigation system in the future.
    
    \item The integration of the depth camera opens opportunities for the development of new algorithms based on machine vision, which can enhance the vehicle's operational capabilities, such as implementing decision trees based on object identification.
    
\end{itemize}

\section{Future work}
Filtering depth points by adjusting the camera's frustum parameters is a simplistic approach. The multilayer costmap can be enhanced by using more sophisticated algorithms, such as density-based point cloud filtering. The current method fails to identify and filter out some single-point data generated either because of camera's noise, or when facing reflective surfaces such as glass. Implementing a density-based filter could potentially mitigate this issue, taking into account that objects have a high point cloud density as can be seen in figure~\ref{fig:costmap_pointcloud_ranges}.

    \bibliographystyle{IEEEtran}
    \bibliography{all_bibliography}
\end{document}